\documentclass{article}

\usepackage{arxiv}

\usepackage[utf8]{inputenc} 
\usepackage[T1]{fontenc}    
\usepackage{hyperref}       
\usepackage{url}            
\usepackage{booktabs}       
\usepackage{amsfonts}       
\usepackage{nicefrac}       
\usepackage{microtype}      
\usepackage{lipsum}

\usepackage{times}
\usepackage{epsfig}
\usepackage{graphicx}
\usepackage{amsmath}
\usepackage{amssymb}
\usepackage{gensymb}
\usepackage{subcaption,graphicx}

\usepackage{xcolor}
\usepackage{cancel}
\usepackage{soul}

\title{Comparison of Spatiotemporal Networks for Learning Video Related Tasks}

\author{
  Logan Courtney \\
  Department of Industrial and Enterprise Systems Engineering\\
  University of Illinois\\
  Urbana-Champaign, IL \\
  \texttt{courtne2@illinois.edu} \\
   \And
 Ramavarapu Sreenivas \\
  Department of Industrial and Enterprise Systems Engineering\\
  University of Illinois\\
  Urbana-Champaign, IL \\
  \texttt{rsree@illinois.edu} \\
}

\begin{document}
\maketitle

\begin{abstract}
Many methods for learning from video sequences involve temporally processing 2D CNN features from the individual frames or directly utilizing 3D convolutions within high-performing 2D CNN architectures. The focus typically remains on how to incorporate the temporal processing within an already stable spatial architecture. This work constructs an MNIST-based video dataset with parameters controlling relevant facets of common video-related tasks: classification, ordering, and speed estimation. Models trained on this dataset are shown to differ in key ways depending on the task and their use of 2D convolutions, 3D convolutions, or convolutional LSTMs. An empirical analysis indicates a complex, interdependent relationship between the spatial and temporal dimensions with design choices having a large impact on a network's ability to learn the appropriate spatiotemporal features.
\end{abstract}


\section{Introduction}
The convolution operation remains at the core of 2D spatial networks due to advantageous characteristics such as weight sharing (reduced parameters), local connectivity (reduced convergence time), and down-sampling (reduced computation). Base architectures have grown deeper and more complex over time and demonstrated improved performance on a variety of image-based tasks \cite{cnn_survey}.

For natural language processing (NLP), techniques for handling 1D sequences differ in how the temporal information is consumed with both convolutions\cite{cnn_sentence_classification} and LSTM networks\cite{lstm} showing early success. LSTMs process elements one at a time with a form of memory as opposed to convolutions operating on local neighborhoods. Although still occasionally used, these methods have largely been replaced by the Transformer\cite{transformer} due to its non-recurrent structure and its ability to self-attend over all positions in the input sequence.

When it comes to video, the success of a deep network depends on its ability to learn both spatial and temporal features. Interpreting a video as a sequence of images, many video-based networks borrow techniques from the above fields of image processing and NLP.

The early stages of action recognition had relatively small datasets\cite{ucf101} insufficient for training novel spatiotemporal networks from scratch. This forced methods to creatively utilize 2D image networks pretrained on ImageNet\cite{imagenet}. \cite{cnn_lstm_ar} stacks LSTMs on top of a pretrained CNN. \cite{two_stream} inputs RGB frames for content and optical flow frames for motion to pretrained CNNs. The spatial and temporal dimensions are handled in two separate stages. As larger action recognition datasets become available such as Kinetics\cite{kinetics1}\cite{kinetics2}\cite{kinetics3}, networks primarily use 3D convolutions to directly capture spatiotemporal features. \cite{3dcnnretrace} uses 3D versions of high-performing 2D CNNs. \cite{fastslow} uses a high-resolution, low frame rate model for spatial semantics and a low-resolution, high frame rate model for motion. \cite{smallbignet} utilizes two spatiotemporal views to exploit different contexts. The focus remains on temporal modifications to popular 2D image networks without an emphasis on how the choice of the spatial architecture directly impacts what temporal modifications are necessary. 

This issue becomes more apparent after viewing recent techniques for lipreading. \cite{resnet_lipreading} demonstrates an increase in performance when utilizing 3D convolutions for only the first layer of the network to process the short-term dynamics. The remainder of the spatial information is processed with 2D convolutions before a bidirectional LSTM to capture long-term context. \cite{convlstm_lipreading} achieves comparable performance utilizing convolutional LSTM layers throughout the full network. Utilizing 3D convolutions throughout the full network in \cite{lrw} results in their lowest performing model indicating a non-negligible relationship between the spatial architecture and temporal technique. Current state-of-the-art lipreading models \cite{asr_is_all_you_need}\cite{watch_to_listen_clearly} both utilize modified training schemes with additional unlabeled data without focusing on the spatiotemporal architecture.

This work takes a step backwards and attempts to understand how various video techniques operate on a more fundamental level. Multiple models are trained on a novel MNIST-based video dataset. By directly parameterizing the video sequence to control for visible spatiotemporal features and training on multiple video related tasks, the performance of these techniques is shown to differ in key ways. Section \ref{section:spatiotemporal} begins with an overview of spatiotemporal receptive fields and resolution. Section \ref{section:dataset} discusses the creation of the MNIST-based video dataset and its associated spatiotemporal tasks. Section \ref{section:models} discusses the various models and their characteristics. Section \ref{section:results} provides an interpretation of the results followed by Section \ref{section:closing_remarks} wrapping up with the impact and future work.

\section{Spatiotemporal Receptive Field and Resolution}\label{section:spatiotemporal}
The Universal Approximation Theorem (UAT) \cite{uat} states a single layer network with a finite number of neurons can approximate any continuous function arbitrarily close. Given a video input $x \in \mathcal{R}^{h\times w\times c \times T}$, a single layer network with $h_o$ hidden units is sufficient for all video related tasks. Each element of the output is dependent on the full spatial and temporal dimensions. In practice, various techniques such as convolutions are used in place of fully connected networks due to the unknown number of hidden units $h_o$ necessary, the large number of parameters, and the unproven ability to learn such functions.

Although the representation power of the network is reduced with the replacement of fully connected layers, this has still led to high-performing models on a wide variety of tasks. Convolutions constrain models to operate only on small, local neighborhoods of the input. The region of the input contributing to the output of a layer is called the receptive field. In order to learn features in the input outside of the receptive field of a single layer, model designers construct deep networks of many layers.

\noindent
\textbf{Spatial Receptive Field:} Spatial pooling and/or strided convolutions are typically used to reduce the height/width of the input frame as it passes deeper into the network. This significantly reduces the computation and greatly extends the spatial receptive field allowing the small localized area of a particular layer to contain more spatially distant information in the original input.

\noindent
\textbf{Spatial Resolution:} This comes at the cost of reducing the spatial resolution. The exact location of small features must be encoded into the depth dimension otherwise this information is lost. The relevance of this depends on the task. Image classification cares less about where the object appears within the image and more about whether or not it appears at all. Object detection requires the precise spatial information to pinpoint the object's location in the original image. Spatial pooling/strided convolutions have been necessary for every deep network. This is done not for the benefit of model representation but out of necessity for modern implementations. The GPU memory and computation requirements become prohibitive if these techniques are not employed due to the large input image dimensions.

\noindent
\textbf{Temporal Receptive Field:} Convolutions treat the temporal dimension in the same way as the spatial dimension. The temporal receptive field grows gradually as the input passes through the network. Only a few neighboring inputs are seen early in the network and long-range information is handled deeper in the network. LSTMs on the other hand maintain an internal state with an additional self connection to pass the information along every timestep. Long-range dependencies can appear at any layer without forcing the network to delay this processing until a later stage.

\noindent
\textbf{Temporal Resolution:} Similar to the spatial dimension, strided convolutions can be used to increase the temporal receptive field at the cost of resolution. A problem like action recognition may employ such techniques since the task is more related to specifying what happens more so than when it happens. The computation is reduced without sacrificing performance since the destruction of the fine-grained temporal information is less relevant to the output. Techniques for lipreading have shown to perform better when the temporal dimension is maintained following the intuition lipreading requires some degree of what happens as well as when it happens. The additional computation and smaller receptive field from maintaining the temporal dimension is a less of a problem than the spatial dimension typically due to most videos having roughly 30 frames per second (much smaller than the height/width of each frame) and the current tasks (like action recognition and lipreading) require a relatively short temporal receptive field of at most a few seconds.

\noindent
\textbf{Spatiotemporal Considerations}
As long as the output has a sufficiently large receptive field, networks seem to perform well. Detecting small objects in a 2D image may use a relatively small spatial receptive field. Full image classification typically uses receptive fields larger than the input image. Classifying single sentences in a document may require the context of a few neighboring sentences while classifying the document may require the full text. 

Most techniques for spatiotemporal related tasks simply borrow and combine the successful techniques for both of these sub-problems. However, individual consideration of the spatial and temporal dimensions remains insufficient for dealing with spatiotemporal features. These techniques result in a complex interaction which is shown here to have a significant impact on performance depending on the nature of the task.

\section{Dataset Construction and Task Description}\label{section:dataset}

\begin{figure}[!ht]
    \centering
    \includegraphics[width=0.45\linewidth]{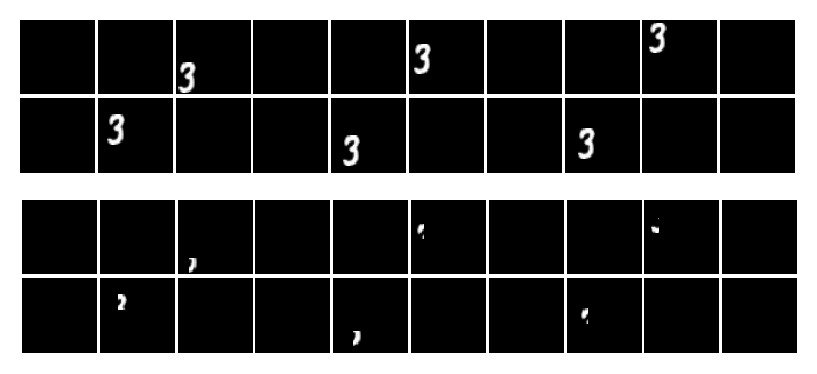}
   \caption{\small Example input from MNIST-based video dataset with and without quadrant masking.}
\label{fig:digit}
\end{figure}

The MNIST dataset \cite{mnist} contains $28 \times 28$ sized images of handwritten digits. These images are used to construct a video dataset (similar to \cite{movingmnist}) meant to test the capability of various networks to learn three spatiotemporal related tasks: classification, ordering, and speed.

The images are placed in a random location of a larger frame of size $64 \times 64$. The digit moves around over the course of 48 frames based on a random pixel speed $S=\{1,2,...,6\}$ and random direction selected at the start of the video. Collisions with walls cause the digit to reverse direction at the frame of impact. Only select frames are visible making the digit appear to blink at a randomly selected rate $V=\{1,2,...,12\}$. The digit is split into four quadrants as shown in the bottom of Figure \ref{fig:digit} with a separate quadrant appearing every visible frame. The four quadrants appear in a randomly selected order with the order repeating itself throughout the video.

\noindent
\textbf{Task \#1 - Digit Classification:} Each network will attempt to identify the digit present in the video. By presenting only a single quadrant every visible frame, the network must piece together information from individual frames. The rate of visible frames $V$ and speed $S$ alter how this information is presented impacting performance based on a network's ability to handle various spatiotemporal features. 2D convolution networks have been shown to achieve high performance classifying the digits providing a reasonable upper bound for expected performance on this task.

\noindent
\textbf{Task \#2 - Ordering:} Each network will attempt to identify the order quadrants appear from six possible sequences: (1,2,3,4), (1,2,4,3), (1,4,2,3), (1,4,3,2), (1,3,2,4), (1,3,4,2). The previous task classifies the digit regardless of the order the information is presented. At the output, the model needs only a summary of the total spatial information. For this task, the spatial information is an intermediate feature allowing the model identify the visible quadrant at each frame while the output only needs to contain the temporal information about the order of presentation.

\noindent
\textbf{Task \#3 - Speed:} Each network will attempt to identify the pixel speed $S$. This task requires the network to maintain information related to both previous tasks as well as maintain high spatial and temporal resolution information. Calculating the speed of the centroid between visible frames is different depending on the quadrant order. Identifying the quadrant order is dependent on identifying the digit.

When $V=12$, a quadrant is shown every 12 frames requiring a minimum of 48 frames for the model to be presented with the full digit. The digit still moves with speed $S$ even if the frame is not visible. $S$ and $V$ are limited to combinations resulting in distances less than 50 pixels between visible frames. That is, at the max speed $S=6$, the max value for $V$ is $8$ (48 total pixels of movement). At a speed $S=4$, $V$ can achieve its max value of $12$. Without this limit, aliasing can occur when the digit bounces off of a wall due to the video frame size of $64 \times 64$.

\section{Models}\label{section:models}
All of the models are based on Residual Networks \cite{resnet}. Each residual block has two paths. The first path contains a convolution with a $1 \times 1$ kernel followed by a specialized convolution (2D convolution, 3D convolution, convolutional LSTM, or 3D convolution with a temporal stride) followed by another 1x1 convolution. The second path is an open connection. Both paths are added together to form the output of the block. If the block uses a spatial stride, the open connection contains a $1 \times 1$ convolution with a stride of 2 to downsample the input to the appropriate size. The models contain 8 total residual blocks which each use a different specialized convolution within the blocks depending on the model. 

\begin{figure}[!ht]
\centering
    \includegraphics[width=0.60\linewidth]{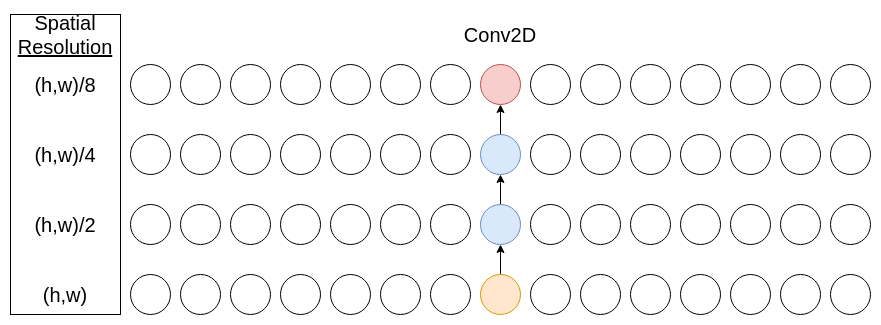}
   \caption{\small The spatial dimension for each frame is processed individually.}
\label{fig:model_conv2d}
\end{figure}

\begin{figure}[!ht]
\centering
    \includegraphics[width=0.60\linewidth]{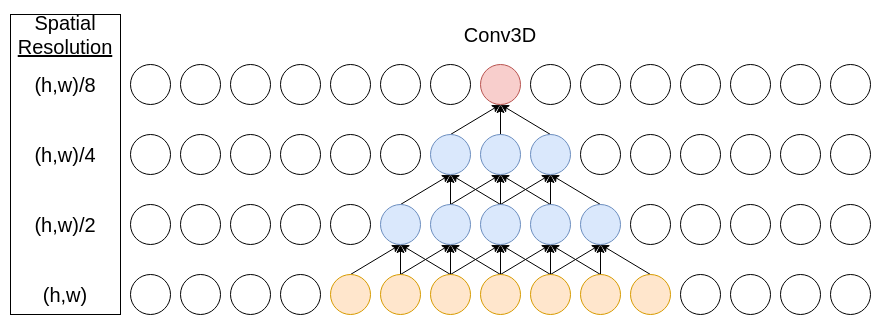}
   \caption{\small The temporal receptive field grows as the network deepens.}
\label{fig:model_conv3d}
\end{figure}

\begin{figure}[!ht]
\centering
    \includegraphics[width=0.60\linewidth]{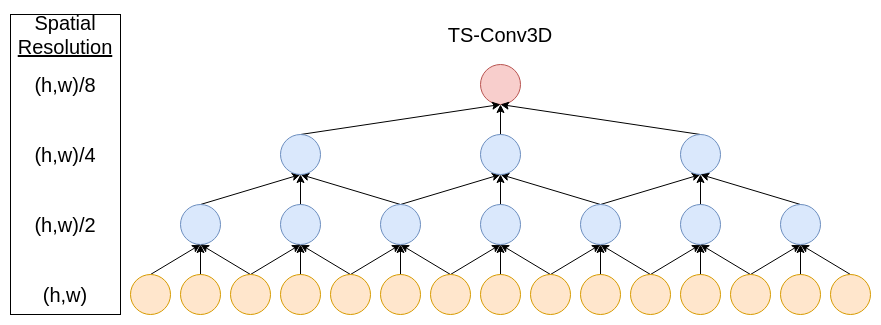}
   \caption{\small The temporal receptive field grows more rapidly as the network deepens due to the temporal stride.}
\label{fig:model_tsconv3d}
\end{figure}

\begin{figure}[!ht]
\centering
    \includegraphics[width=0.60\linewidth]{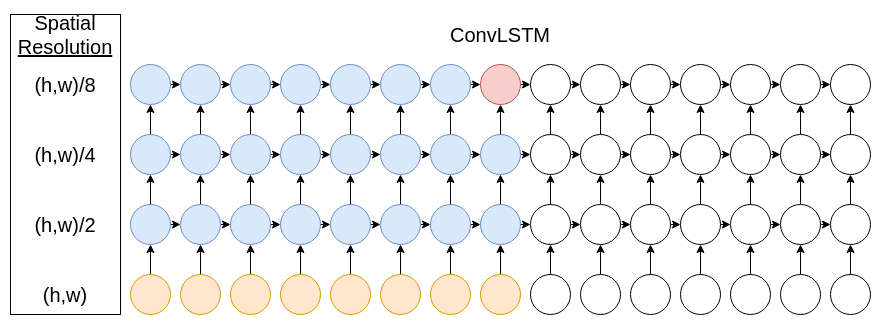}
   \caption{\small All previous inputs are used in calculating the output.}
\label{fig:model_convlstm}
\end{figure}

Tables \ref{table:model_conv2d}, \ref{table:model_conv3d}, \ref{table:model_tsconv3d}, and \ref{table:model_convlstm} specify the number of channels, the convolution stride, the input/output dimensions, and the spatial/temporal receptive fields (SRF/TRF). There are two models of different size for each type. The specialized convolutions operate on the temporal dimension in various ways shown in Figures \ref{fig:model_conv2d}, \ref{fig:model_conv3d}, \ref{fig:model_tsconv3d}, and \ref{fig:model_convlstm}. Each output (red) represents a spatial feature map with the connections highlighting its dependencies on the intermediate feature maps (blue) and input frames (yellow). Unlike 1D temporal problems, the effect of the spatial dimension complicates the impact of these temporal differences since the spatial resolution and receptive field differ throughout the network. 
\subsection{2D Convolution with LSTM (Conv2D)}

\begin{table}[!ht]
\centering
\begin{tabular}{|l|c|c|c|c|c|c|c|}
\hline
Conv2D         & Channels     & Channels-XL  & Stride  & Input Size & Output Size & SRF     & TRF \\ \hline
               & (in,ch,out)  & (in,ch,out)  & (h,w,t) & (h,w,t)    & (h,w,t)     & (h,w)   & (t) \\ \hline
Res1\_1        & (1,16,64)    & (1,36,72)    & (2,2,1) & 64x64x48   & 32x32x48    & 3x3     & 1   \\ \hline
Res2\_1        & (64,16,64)   & (72,36,72)   & (2,2,1) & 32x32x48   & 16x16x48    & 7x7     & 1   \\ \hline
Res3\_1        & (64,16,64)   & (72,36,72)   & (2,2,1) & 16x16x48   & 8x8x48      & 15x15   & 1   \\ \hline
Res3\_2        & (64,16,64)   & (72,36,72)   & (1,1,1) & 8x8x48     & 8x8x48      & 31x31   & 1   \\ \hline
Res4\_1        & (64,32,128)  & (72,72,144)  & (2,2,1) & 8x8x48     & 4x4x48      & 47x47   & 1   \\ \hline
Res4\_2        & (128,32,128) & (144,72,144) & (1,1,1) & 4x4x48     & 4x4x48      & 79x79   & 1   \\ \hline
Res4\_3        & (128,32,128) & (144,72,144) & (1,1,1) & 4x4x48     & 4x4x48      & 111x111 & 1   \\ \hline
Res4\_4        & (128,32,128) & (144,72,144) & (1,1,1) & 4x4x48     & 4x4x48      & 143x143 & 1   \\ \hline
SpatialPool & -            & -            & (4,4,1) & 4x4x48     & 1x1x48      & -       & -   \\ \hline
LSTM1          & (128,128)    & (144,144)    & -       & 1x1x48     & 1x1x48      & -       & 48  \\ \hline
LSTM2          & (128,128)    & (144,144)    & -       & 1x1x48     & 1x1x48      & -       & 48  \\ \hline
\end{tabular}
\caption{\small The temporal receptive field (TRF) remains 1 until the LSTM layers.}
\label{table:model_conv2d}
\end{table}
A baseline network utilizes 2D convolutions followed by two LSTM layers. The spatial and temporal information is handled in two separate stages. The spatial information for a single frame is processed entirely with the spatial dimension fully collapsed before any sort of temporal processing is performed by the LSTM. This means the embedding output from the 2D convolutions must encode the relevant spatial information into the depth dimension.

Consider two sub-functions. The first function takes the input frame and outputs a two-dimensional vector containing the exact $x$ and $y$ pixel location of the digit. The second function calculates the distance between points divided by the number of frames between these points. Combining these functions would result in perfect speed prediction. It is not unreasonable to believe a 2D convolution network and an LSTM could learn these two respective functions suggesting more advanced techniques are unneeded. However, this function may not be learnable via gradient descent with a reasonable network size.

\subsection{3D Convolution with LSTM (Conv3D)}

\begin{table}[!ht]
\centering
\begin{tabular}{|l|c|c|c|c|c|c|c|}
\hline
Conv3D         & Channels     & Channels-XL  & Stride  & Input Size & Output Size & SRF     & TRF \\ \hline
               & (in,ch,out)  & (in,ch,out)  & (h,w,t) & (h,w,t)    & (h,w,t)     & (h,w)   & (t) \\ \hline
Res1\_1        & (1,16,64)    & (1,36,72)    & (2,2,1) & 64x64x48   & 32x32x48    & 3x3     & 3   \\ \hline
Res2\_1        & (64,16,64)   & (72,36,72)   & (2,2,1) & 32x32x48   & 16x16x48    & 7x7     & 5   \\ \hline
Res3\_1        & (64,16,64)   & (72,36,72)   & (2,2,1) & 16x16x48   & 8x8x48      & 15x15   & 7   \\ \hline
Res3\_2        & (64,16,64)   & (72,36,72)   & (1,1,1) & 8x8x48     & 8x8x48      & 31x31   & 9   \\ \hline
Res4\_1        & (64,32,128)  & (72,72,144)  & (2,2,1) & 8x8x48     & 4x4x48      & 47x47   & 11  \\ \hline
Res4\_2        & (128,32,128) & (144,72,144) & (1,1,1) & 4x4x48     & 4x4x48      & 79x79   & 13  \\ \hline
Res4\_3        & (128,32,128) & (144,72,144) & (1,1,1) & 4x4x48     & 4x4x48      & 111x111 & 15  \\ \hline
Res4\_4        & (128,32,128) & (144,72,144) & (1,1,1) & 4x4x48     & 4x4x48      & 143x143 & 17  \\ \hline
SpatialPool & -            & -            & (4,4,1) & 4x4x48     & 1x1x48      & -       & -   \\ \hline
LSTM1          & (128,128)    & (144,144)    & -       & 1x1x48     & 1x1x48      & -       & 48  \\ \hline
LSTM2          & (128,128)    & (144,144)    & -       & 1x1x48     & 1x1x48      & -       & 48  \\ \hline
\end{tabular}
\caption{\small The temporal receptive field does not grow fast enough to cover the full sequence of length 48 before the spatial dimension collapses.}
\label{table:model_conv3d}
\end{table}

A deep network of 3D convolutions, which has two LSTM-layers at the top as before, processes the temporal and spatial information simultaneously. For short term dynamics, the 3D convolution is able to directly extract spatiotemporal features if they fall within its spatiotemporal receptive field. For long term dynamics, the model must encode the high-resolution spatial information into the depth dimension to be processed by a later layer. This is similar to the 2D convolution network but less extreme as it can still handle some degree of temporal information. For example, speeds $S\geq 3$ move the object outside of the spatial receptive field of the first layer with visible frame rates $V\geq3$ moving the object outside of its temporal receptive field. For these large values, the model is forced to combine information from multiple frames much later in the network at a reduced spatial resolution.

\subsection{Temporally Strided 3D Convolutions with LSTM (TS-Conv3D)}\label{subsection:tsconv3d}

\begin{table}[!ht]
\centering
\begin{tabular}{|l|c|c|c|c|c|c|c|}
\hline
TS-Conv3D-XL   & Channels    & Channels-XL  & Stride  & Input Size & Output Size & SRF     & TRF \\ \hline
               & (in,ch,out) & (in,ch,out)  & (h,w,t) & (h,w,t)    & (h,w,t)     & (h,w)   & (t) \\ \hline
Res1\_1        & -           & (1,36,72)    & (2,2,2) & 64x64x48   & 32x32x24    & 3x3     & 3   \\ \hline
Res2\_1        & -           & (72,36,72)   & (2,2,2) & 32x32x24   & 16x16x12    & 7x7     & 7   \\ \hline
Res3\_1        & -           & (72,36,72)   & (2,2,2) & 16x16x12   & 8x8x6       & 15x15   & 15  \\ \hline
Res3\_2        & -           & (72,36,72)   & (1,1,1) & 8x8x6      & 8x8x6       & 31x31   & 31  \\ \hline
Res4\_1        & -           & (72,72,144)  & (2,2,2) & 8x8x3      & 4x4x3       & 47x47   & 47  \\ \hline
Res4\_2        & -           & (144,72,144) & (1,1,1) & 4x4x3      & 4x4x3       & 79x79   & 79  \\ \hline
Res4\_3        & -           & (144,72,144) & (1,1,1) & 4x4x3      & 4x4x3       & 111x111 & 111 \\ \hline
Res4\_4        & -           & (144,72,144) & (1,1,1) & 4x4x3      & 4x4x3       & 143x143 & 143 \\ \hline
SpatialPool & -           & -            & (4,4,1) & 4x4x3      & 1x1x3       &         &     \\ \hline
LSTM1          & -           & (144,144)    & -       & 1x1x3      & 1x1x3       & -       & 48  \\ \hline
LSTM2          & -           & (144,144)    & -       & 1x1x3      & 1x1x3       & -       & 48  \\ \hline
\end{tabular}
\caption{\small The temporal receptive field (TRF) grows at the same rate as the spatial receptive field (SRF).}
\label{table:model_tsconv3d}
\end{table}
The previous model is modified to use strided convolutions in the temporal dimension. The temporal receptive field increases more rapidly at the cost of decreased temporal resolution in the same manner as the spatial resolution. For short term dynamics, the network is capable of dealing with spatiotemporal features sooner than the no-stride network but now must also encode the temporal location into the depth dimension. This can be both a benefit and a hindrance. A particular layer has a wider field of vision allowing it to directly act on the relevant information before the spatial dimension gets reduced but the model is directly acting on less precise information due to the temporal resolution.

\subsection{Convolutional LSTM (ConvLSTM)}

\begin{table}[!ht]
\centering
\begin{tabular}{|l|c|c|c|c|c|c|c|}
\hline
ConvLSTM       & Channels     & Channels-XL  & Stride  & Input Size & Output Size & SRF     & TRF \\ \hline
               & (in,ch,out)  & (in,ch,out)  & (h,w,t) & (h,w,t)    & (h,w,t)     & (h,w)   & (t) \\ \hline
Res1\_1        & (1,16,64)    & (1,16,64)    & (2,2,1) & 64x64x48   & 32x32x48    & 3x3     & 48  \\ \hline
Res2\_1        & (64,16,64)   & (64,16,64)   & (2,2,1) & 32x32x48   & 16x16x48    & 7x7     & 48  \\ \hline
Res3\_1        & (64,16,64)   & (64,16,64)   & (2,2,1) & 16x16x48   & 8x8x48      & 15x15   & 48  \\ \hline
Res3\_2        & (64,16,64)   & (64,16,64)   & (1,1,1) & 8x8x48     & 8x8x48      & 31x31   & 48  \\ \hline
Res4\_1        & (64,32,128)  & (64,32,128)  & (2,2,1) & 8x8x48     & 4x4x48      & 47x47   & 48  \\ \hline
Res4\_2        & (128,32,128) & (128,32,128) & (1,1,1) & 4x4x48     & 4x4x48      & 79x79   & 48  \\ \hline
Res4\_3        & (128,32,128) & (128,32,128) & (1,1,1) & 4x4x48     & 4x4x48      & 111x111 & 48  \\ \hline
Res4\_4        & (128,32,128) & (128,32,128) & (1,1,1) & 4x4x48     & 4x4x48      & 143x143 & 48  \\ \hline
SpatialPool & -            & -            & (4,4,1) & 4x4x48     & 1x1x48      & -       & -   \\ \hline
\end{tabular}
\caption{\small The temporal receptive field (TRF) is large at all layers of the network.}
\label{table:model_convlstm}
\end{table}
A deep network of convolutional LSTMs the spatiotemporal features simultaneously. The hidden state for each layer is passed to each subsequent frame implying each layer is capable of processing any temporal feature at all stages of spatial resolution and spatial receptive field. As the spatial resolution is decreased, the model can either encode the spatial information into the depth dimension or maintain this information in the hidden state to be processed at the next time step. Each layer having access to the full temporal information regardless of the network depth is an expected advantage.

\section{Results}\label{section:results}

\begin{table}[!ht]
\centering
\begin{tabular}{|l|c|c|c|c|c|}
\hline
 & Task \#1     & Task \#2        & Task \#3        & Number of      & Total Training            \\ \hline
             & Classification  & Ordering        & Speed          & Parameters     & Time       \\ \hline
Conv2D       & \textbf{97.9\%} & \textbf{98.5\%} & 0.166          & 385K           & 11.1 hours \\ \hline
Conv3D       & \textbf{97.9\%} & \textbf{98.5\%} & 0.187          & 473K           & 11.7 hours \\ \hline
ConvLSTM     & 97.2\%          & 98.1\%          & \textbf{0.129} & 437K           & 15.2 hours \\ \hline
             &                 &                 &                &                &            \\ \hline
Conv2D-XL    & \textbf{98.5\%} & \textbf{98.9\%} & 0.164          & 702K           & 12.5 hours \\ \hline
Conv3D-XL    & 98.3\%          & \textbf{98.9\%} & 0.147          & 1,145K         & 17.8 hours \\ \hline
TS-Conv3D-XL & 96.5\%          & 92.0\%          & 0.146          & 1,145K         & 8.1 hours  \\ \hline
ConvLSTM-XL  & 97.8\%          & 98.8\%          & \textbf{0.059} & 1,967K         & 28.3 hours \\ \hline
\end{tabular}
\caption{\small Performance over the three tasks for all models trained. The metric for the third task is mean absolute error (MAE) of the pixel speed.}
\label{table:results}
\end{table}

\subsection{Digit Classification and Sequence Order}
As shown in Table \ref{table:results}, there were slight variations in performance for the first two tasks. The large models all outperformed their smaller counterparts. With a magnitude of less than 1\%, both convolutional LSTM models lagged behind the 2D/3D models digit classification.

Excluding TS-Conv3D-XL, performance was largely unaffected by the input parameters. Speed S had no effect on either task. Large values for visible frames V caused a drop between 2\% and 3\% when classifying the sequence order compared to smaller values. The models have less information considering each quadrant is visible only once in these scenarios.

The TS-Conv3D-XL model is the only model to stand out. The drastic drop in performance for classifying the sequence order is almost entirely due to when $V=1$ (70.8\%) and $V=2$ (85.4\%). Performance for $V\geq3$ matches closely with Conv3D-XL. This difference must be attributed to the decrease in temporal resolution early in the network. For small values of V, there is no longer a unique temporal input for each individual quadrant after the first layer. The model is forced to encode this information into the depth dimension as opposed to the Conv3D-XL model can delay this until later in the network.

This decrease in sequence order performance is most likely connected to the under-performance in digit classification. Unless a digit is recognizable from only 1-2 quadrants (such as the number 1), the model must either know the sequence order to appropriately piece together the parts or know the digit in order to identify the order. Digits 6, 8, and 9 have higher error rates.  Three out of four quadrants of the digit 8 match with 6 or 9. For these small values of V, the model immediately had to encode two separate frames into a single output making this distinction later in the network difficult. This difference in digit error rates is only seen in TS-Conv3D-XL.

\subsection{Speed Regression}
Performance estimating the speed is far more varied between the models. Mean absolute error is used as the metric. 3D convolutions outperformed 2D convolutions and convolutional LSTMs outperformed 3D convolutions. At a first glance, this is rather uninsightful considering the difference in spatiotemporal receptive fields between these models was already known ahead of time. However, unlike the first two tasks, performance is heavily impacted by the parameters $S$ and $V$. By comparing performance over the space of $S$ and $V$, the variations provide a insight into how these techniques operate.

\begin{figure}[!ht]
    \centering
    \begin{subfigure}[!ht]{0.33\textwidth}
        \centering
        \includegraphics[width=\textwidth]{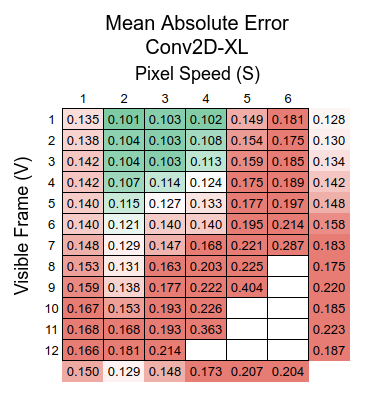}
        \label{fig:mae_conv2d}
    \end{subfigure}
    \vspace{-2\baselineskip}
    \begin{subfigure}[!ht]{0.33\textwidth}  
        \centering 
        \includegraphics[width=\textwidth]{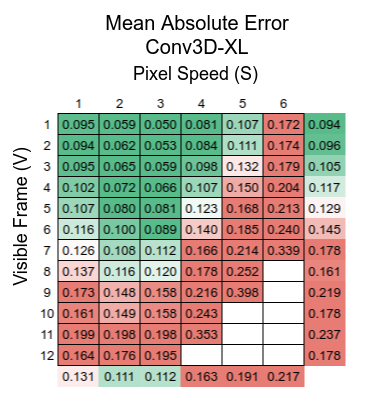}
        \label{fig:mae_conv3d}
    \end{subfigure}
    \begin{subfigure}[!ht]{0.33\textwidth}   
        \centering 
        \includegraphics[width=\textwidth]{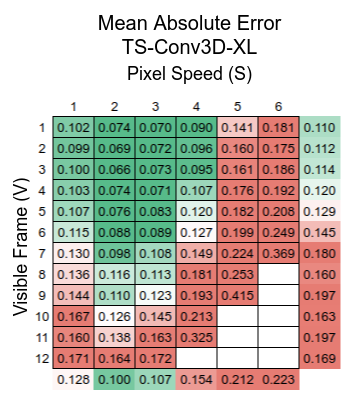}
        \label{fig:mae_tsconv3d}
    \end{subfigure}
    \begin{subfigure}[!ht]{0.33\textwidth}   
        \centering 
        \includegraphics[width=\textwidth]{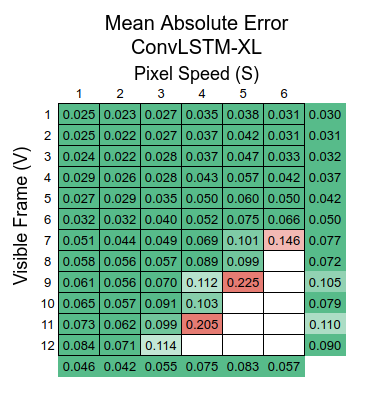}
        \label{fig:mae_convlstm}
    \end{subfigure}
    \caption[]
    {\small Mean absolute error (MAE) over the space of V and S for all four models.} 
    \label{fig:mae_speed}
\end{figure}

\begin{figure}[!ht]
\centering
    \includegraphics[width=0.35\linewidth]{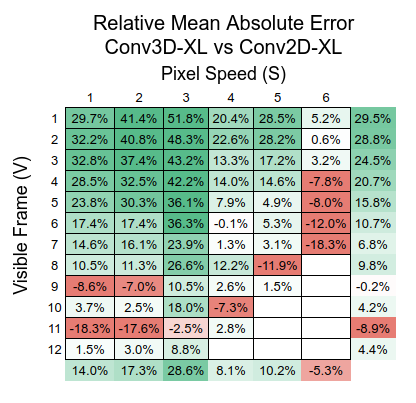}
   \caption{\small The Conv3D model performs significantly better than the Conv2D model for small values of V and S yet performs poorly on its spatiotemporal boundaries. }
\label{fig:mae_conv2d_conv3d}
\end{figure}

\noindent
\textbf{Conv2D-XL vs. Conv3D-XL:} Although Conv3D-XL has a 10\% smaller overall MAE than Conv2D-XL, the error difference is non-uniform over the space of $S$ and $V$ with the relative performance between the models shown in Figure \ref{fig:mae_conv2d_conv3d}. Positive values indicate greater performance by Conv3D. For slow speeds ($S\leq3$) and small values of V ($\leq5$), the inputs fall into the spatiotemporal receptive field of the first three residual blocks allowing the model to actually take advantage of the 3D convolution operation and significantly outperform Conv2D. The Conv2D model must instead correctly encode this spatial information for later processing by the LSTM layers.

For $V\geq9$, performance is nearly identical between the models. Note the temporal receptive field for Conv3D is 9 after Res3\_2 (see Table \ref{table:model_conv3d}). These large values of V prevent the model from any temporal operations until after this layer implying it operates like a 2D convolution. The spatial resolution has decreased by a factor of 8 at this point making this encoding significantly more difficult resulting in a sharp drop in MAE.

The region $4\leq V\leq 7$ and $S=6$ is on the cusp of the spatiotemporal receptive field for Conv3D. Between visible frames, the object moves between 24 and 42 pixels. Although the input falls within the temporal receptive field of Res3, the input is occasionally outside of the spatial receptive field causing the model to delay processing until after the spatial resolution has been reduced. These regions where the network operates as a hybrid of 3D convolutions and 2D convolutions seem to cause learning difficulties resulting in performance worse than the model with only 2D convolutions.

\begin{figure}[!ht]
\centering
    \includegraphics[width=0.35\linewidth]{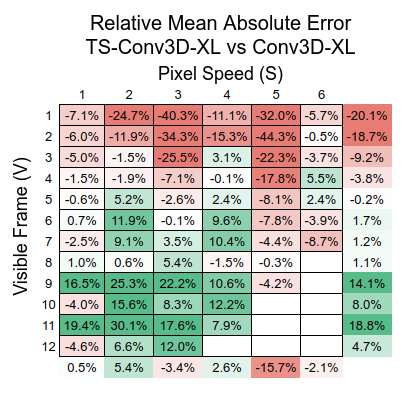}
   \caption{\small Both Conv3D models achieve the same MAE over the entire dataset but very different errors over the space of V and S.}
\label{fig:mae_conv3d_tsconv3d}
\end{figure}

\noindent
\textbf{Conv3D-XL vs. TS-Conv3D-XL:} Both models achieved comparable overall MAE with the relative differences over the space of $S$ and $V$ shown in Figure \ref{fig:mae_conv3d_tsconv3d}. The temporal receptive field increases more rapidly allowing the model access to spatial information before the spatial resolution has decreased at the cost of decreased temporal resolution.

For $V\geq 9$, the input now falls into the temporal receptive field of TS-Conv3D before the spatial resolution has reduced resulting in an increase of performance over the Conv3D model. For $V\leq 4$, the input falls into the temporal receptive field of both models at a high spatial resolution yet with reduced temporal resolution for TS-Conv3D hindering performance. Temporally strided convolutions can be both a benefit and a hindrance.

\begin{figure}[!ht]
\centering
    \includegraphics[width=0.35\linewidth]{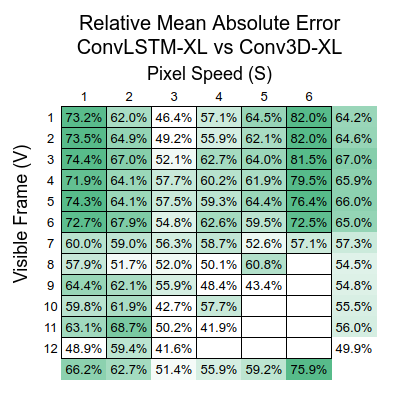}
   \caption{\small The ConvLSTM model performs relatively better than the Conv3D model over the entire space of V and S.}
\label{fig:mae_convlstm_conv3d}
\end{figure}

\noindent
\textbf{Conv3D-XL vs. ConvLSTM-XL:} The relative performance difference between Conv3D-XL and ConvLSTM-XL over the space of $S$ and $V$ is shown in Figure \ref{fig:mae_convlstm_conv3d}. ConvLSTM performed better for all inputs. Allowing each layer access to the full range of temporal information provides a clear advantage for estimating the speed. Additionally, the performance is far more uniform without the model being biased towards particular regions unlike the Conv3D and TS-Conv3D models.

Although the use of convolutional LSTMs increases the temporal receptive field over 3D convolutions, it does not compensate for the full spatiotemporal receptive field. Performance still gradually drops as both $S$ and $V$ increase. Large values of these parameters result in very large object movements between visible frames. Convolutional LSTMs still utilize 2D convolutions for spatial operations and these large movements can easily fall outside the spatial receptive field delaying processing until a later stage after the spatial resolution has been reduced. 

\section{Closing Remarks}\label{section:closing_remarks}

\noindent
\textbf{Encoding Spatial Information:} When the input falls outside of the spatiotemporal receptive field of a particular 3D convolution layer, the network simply operates as a 2D convolution and encodes the spatial information into the depth dimension for a later stage. Although Conv2D performed the worst for estimating the speed, it works unexpectedly well with 96.1\% accuracy when the output is rounded. Encoding the spatial information is clearly doable but may be difficult to learn via gradient descent or may simply require a prohibitively large model. The models with a large depth dimension all outperformed their smaller counterparts yet still underperformed relative to the smaller convolutional LSTM model. Utilizing additional connections to allow access to as much of the spatiotemporal information as possible at all stages of processing is more effective than forcing the model to learn in a roundabout way such as forced encoding.

\noindent
\textbf{Differences Between Spatial and Temporal Resolution:} The underperformance of temporally strided 3D convolutions for digit classification and sequence order hints at the importance of maintaining the temporal dimension. Is this also true for the spatial dimension? Nearly all 2D image networks reduce the spatial resolution out of necessity. Without doing so would require very deep networks to achieve a large visual receptive field. This very deep network would also be operating on much higher resolution feature maps resulting in a prohibitive amount of computation. If the information in these dimensions is different, this may not be an issue. However, if they are different, is there a reason for treating the dimensions the same as is done by 3D convolutions?

\noindent
\textbf{Effective Use of 3D Convolutions:} Performance of 3D convolutions is heavily impacted by design choices. Conv3D and TS-Conv3D appear equal overall with this particular dataset but that is simply by design. Both models learn a different spatiotemporal region. For real datasets, the distribution of spatiotemporal features is almost certainly non-uniform. This is the crux of the problem. Should the network be well designed to capture the appropriate spatiotemporal features, performance will be high. However, the scale of the important spatiotemporal features is frequently unknown ahead of time requiring extensive experimentation.

This is further complicated after comparing the results of Conv2D-XL and Conv3D-XL. When an input falls on the spatiotemporal boundary, the 3D model can actually perform worse. A poorly designed 3D network could lead to more problems and create difficulties identifying the source of errors during this experimentation stage.

\noindent
\textbf{Overfitting with Convolutional LSTMs:} Convolutional LSTMs seem capable of learning a much larger spatiotemporal region than 3D convolutions. This also provides a greater capability to overfit. Without the ability to parameterize the spatiotemporal receptive field like 3D convolutions, it is not possible to directly control which spatiotemporal features the model focuses on. Without a significant amount of data for the model to learn which features to focus on, it could easily overfit to regions of the feature space with a smaller signal to noise ratio. 3D convolutions inherently filter information which may explain their relatively high performance in a lot of real world datasets.

\noindent
\textbf{Impact of Task and Loss Function:} Although the greater capabilities of convolutional LSTMs were necessary for a task like speed estimation, the model underperformed for digit classification. There is a difference between learning a problem requiring spatiotemporal features and a temporal problem involving spatial features. Digit classification simply needs to know what happens and not when it happens. Although convolutional LSTMs have greater spatiotemporal capabilities, they may not learn those capabilities as easily as 3D convolutions. Difficulties in training LSTMs compared with convolutions has been frequently discussed in other fields like NLP.

\noindent
\textbf{Limitation of Experiments:} These drastic performance differences appear even in this relatively simple MNIST-based video dataset. The digits are in motion yet they are rigid with no rotation. They are 2D flat objects yet most videos contain 2D projections of 3D objects. It is easy to imagine many real datasets containing much more complex spatiotemporal frequencies along with large varying scales, camera movement, color, and motion blur.

{\small
\bibliographystyle{unsrt}
\bibliography{template}
}

\end{document}